\documentclass[letterpaper, 10 pt, conference]{ieeeconf}  % Comment this line out if you need a4paper

\IEEEoverridecommandlockouts	% to override locked commands
\usepackage{tikz}
\usepackage{tabularx}
\usepackage{amsmath}
\usepackage{adjustbox}
\usepackage{dirtree}
\usepackage[T1]{fontenc}
\usepackage{geometry}
\usepackage[normalem]{ulem}

\usepackage{geometry}
\def\checkmark{\tikz\fill[scale=0.4](0,.35) -- (.25,0) -- (1,.7) -- (.25,.15) -- cycle;}

\usepackage{array,booktabs,arydshln,xcolor}

\makeatletter % changes the catcode of @ to 11
\newcommand{\linebreakand}{%
  \end{@IEEEauthorhalign}
  \hfill\mbox{}\par
  \mbox{}\hfill\begin{@IEEEauthorhalign}
}
\title{\LARGE \bf Automated Vehicle Driver Monitoring Dataset from Real-World Scenarios}

\author{ Mohamed Sabry$^{1}$  \emph{Student Member, IEEE}, Walter Morales-Alvarez$^{1}$ \emph{Student Member, IEEE}, \linebreakand and Cristina Olaverri-Monreal$^{1}$ \emph{Senior Member, IEEE}
    \thanks{$^{1}$  Johannes Kepler University Linz, Austria, Department Intelligent
Transport Systems, Altenberger Straße 69, 4040 Linz, Austria.
        {\tt\small \{mohamed.sabry,  walter.morales$\_$alvarez, cristina.olaverri-monreal\}@jku.at}}
}

\usepackage{atbegshi}
\usepackage[absolute,overlay]{textpos}
\usepackage{rotating}

\usepackage[english]{babel}
\usepackage{blindtext}

\usepackage{framed}

\begin{document}

\maketitle
\thispagestyle{empty}
\pagestyle{empty}
\setlength{\textfloatsep}{1\baselineskip plus 0.2\baselineskip minus 0.2\baselineskip}

\setlength{\dbltextfloatsep}{1\baselineskip plus 0.2\baselineskip minus 0.2\baselineskip}

%%%%%%%%%%%%%%%%%%%%%%%%%%%%%%%%%%%%%%%%%%%%%%%%%%%%%%%%%%%%%%%%%%%%%%%%%%%%%%%%
\begin{abstract}

From SAE Level 3 of automation onwards, drivers are allowed to engage in activities that are not directly related to driving during their travel. However, in level 3, a misunderstanding of the capabilities of the system might lead drivers to engage in secondary tasks, which could impair their ability to react to challenging traffic situations.

Anticipating driver activity allows for early detection of risky behaviors, to prevent accidents.
To be able to predict the driver activity, a Deep Learning network needs to be trained on a dataset. However, the use of datasets based on simulation for training and the migration to real-world data for prediction has proven to be suboptimal. Hence, this paper presents a real-world driver activity dataset, openly accessible on IEEE Dataport, which encompasses various activities that occur in autonomous driving scenarios under various illumination and weather conditions. Results from the training process showed that the dataset provides an excellent benchmark for implementing models for driver activity recognition.
% showed an accurate prediction of the in-vehicle actions performed. 

\end{abstract}

\begin{textblock*}{18.15cm}(1.55cm,26cm) % Adjust width and position as needed
\begin{minipage}{17.8cm}
     \vspace{0.1cm} % Vertical space within the minipage
     {\footnotesize\copyright 2024 IEEE. Personal use of this material is permitted. Permission from IEEE must be obtained for all other uses, in any current or future media, including reprinting/republishing this material for advertising or promotional purposes, creating new collective works, for resale or redistribution to servers or lists, or reuse of any copyrighted component of this work in other works. DOI: 10.1109/ITSC58415.2024.10920048}
\end{minipage}
\end{textblock*}

%%%%%%%%%%%%%%%%%%%%%%%%%%%%%%%%%%%%%%%%%%%%%%%%%%%%%%%%%%%%%%%%%%%%%%%%%%%%%%%%
\section{Introduction}
\label{sec:introduction}

Automated driving (AD) aims to reduce driver workload and enhance safety. The Society of Automotive Engineers (SAE) categorizes vehicle automation levels based on system and driver responsibilities~\cite{saelevels}. Most commercial vehicles currently operate at SAE Level 2, where driver assistance systems are present, but the driver remains primarily responsible for driving tasks.

At Level 3 automation, the system assumes the primary driving role under specific conditions, with the driver acting as a fallback in emergencies when the system reaches its operational design domain. However, drivers may overestimate the system's capabilities at this level, engaging in secondary activities and failing to respond promptly in critical situations~\cite{walter_review_2020}.
In scenarios like platooning, understanding preceding activities is crucial for assessing risks and determining necessary alerts before platoon disconnection. Given these challenges, developing Deep Learning (DL) systems to monitor driver activities and provide timely alerts is essential for ensuring safe transitions between automated and manual driving modes.

To optimize the performance of DL driver activity classification models, it's essential to utilize a dataset that encompasses a diverse array of real-world scenarios. This ensures that the training data is representative of actual driving conditions, providing crucial information for the network to effectively generalize across varied driving situations, especially in image-based models. This entails incorporating diverse weather and lighting conditions, a dynamic background (such as moving vehicles), and authentic passenger reactions.

Existing driver monitoring datasets, such as those published in \cite{tran2020real} \cite{katrolia2021ticam} \cite{kopuklu2021driver}, primarily rely on non-naturalistic data collected in simulators, lacking crucial real-world elements, such as varying illumination. Consequently, DL models trained on these datasets perform poorly and generalize inadequately during real-life testing \cite{morales2023transferability}.

To address the lack of real-world driver monitoring datasets with dynamic automated driving capabilities, we present the  Johannes Kepler University-Intelligent Transport Systems (JKU-ITS) Automated Vehicle Driver Monitoring (AVDM) dataset.

The JKU-ITS AVDM dataset comprises data from 17 participants performing tasks with varying levels of distraction. Data collection adhered to relevant guidelines and regulations, with informed consent obtained from all participants. The dataset is openly accessible on IEEE Dataport at~\cite{AVDM}.

Data was collected using the JKU-ITS research vehicle, which possesses automated capabilities~\cite{certad2022jku}. The experiments were conducted under diverse illumination and weather conditions along a secure test route within the Johannes Kepler University (JKU) campus. 

This dataset aims to serve as training data for models designed to recognize non-driving-related activities in SAE Level 3 and higher automated vehicles, contributing to accident prevention efforts. To establish a benchmark, we trained a baseline model for driver activity detection using this dataset. Specifically, we trained and evaluated the Inflated 3D (I3D) architecture~\cite{carreira2017quo} on the proposed dataset, providing a reference point for future work.

This paper is structured as follows: Section II reviews previous work, Section III provides a detailed description of the proposed dataset, and presents an overview of the I3D DL network used. Section IV details the results of the I3D on the proposed dataset, including the experimental setup and discussion. Finally, Section V summarizes the results and suggests potential areas for future research.

\section{Related Work}
\label{sec:RelatedWork}

Several datasets of in-vehicle activities in real traffic scenarios have been created over the last 10 years. They contain multiple actions that drivers engage in while manually driving a car.
% , as illustrated in Figure \ref{fig:postureActivity}. 
These datasets include data such as steering wheel movements, pedal usage, and gaze direction, etc. \cite{yan2016driving}. Similarly, studies such as \cite{abouelnaga2017real} investigated the significance of hand and facial movements in detecting and categorizing driver distractions. This led to the creation of a dataset specifically focusing on the distracted driver posture.
Other studies have proposed a multi-view, multimodal camera-based framework for recognizing driver activities, leveraging information from the head, eyes, and hands position to provide a comprehensive understanding of the behaviour of the driver \cite{ohn2014head}, \cite{jOrtega2020} and for enhancing human-machine interface applications within vehicles \cite{ohn2014hand}. Another direction was to fuse camera and Radar data for the driver activity recognition task in a simulated environment \cite{Li2024ADA}.

% \textcolor{red}{
However, the aforementioned approaches are limited to examining the interaction between the driver and the vehicle in traditional driving scenarios (SAE level 2 autonomy and below), where the driver actively participates in driving tasks. These activities include engagement with the steering wheel, gear lever, and display panel. As a result, these methodologies do not adequately address the tasks that drivers may engage in within the context of automated driving (SAE level 3 or higher) and which could influence their readiness and time to handle a Take Over Request (TOR) \cite{winzer2017modifications}.

In simulated environments, numerous studies have explored TORs through data analysis (e.g., \cite{morales2022automated}, \cite{gold2013take}). However, according to \cite{morales2021real} and \cite{morales2020vehicle}, investigations into real-world driving scenarios remain uncommon.

Similarly, existing datasets for classifying driver activities in the context of AD are predominantly located in simulated environments. 

A comprehensive collection of data can be found in the Drive and Act dataset~\cite{martin2019drive}, which includes detailed recordings of various driver activities such as drinking, eating, reading etc., all captured within simulated automated driving scenarios.

The dataset provides multiple camera views of the driver, depth and skeleton information, among other aspects. However, the data were collected within a stationary vehicle inside a laboratory with a simulated environment, leading to noticeably more relaxed participant behavior compared to situations involving moving vehicles with automated driving capabilities. Furthermore, the dataset lacks real-world variations in illumination and weather conditions.

Being in a moving vehicle requires drivers to remain attentive to their surroundings to quickly regain control in the event of a sudden situation that could lead to accidents. 

There is a notable absence of comprehensive datasets focusing on activities conducted in the context of automated driving under real-world conditions.
To address this gap, this paper presents, what, to the best of the authors' knowledge, is the first driver activity monitoring dataset collected in a real vehicle environment \cite{certad2022jku}. 

\begin{figure}[]
  \centering
  \includegraphics[width=0.55\columnwidth]{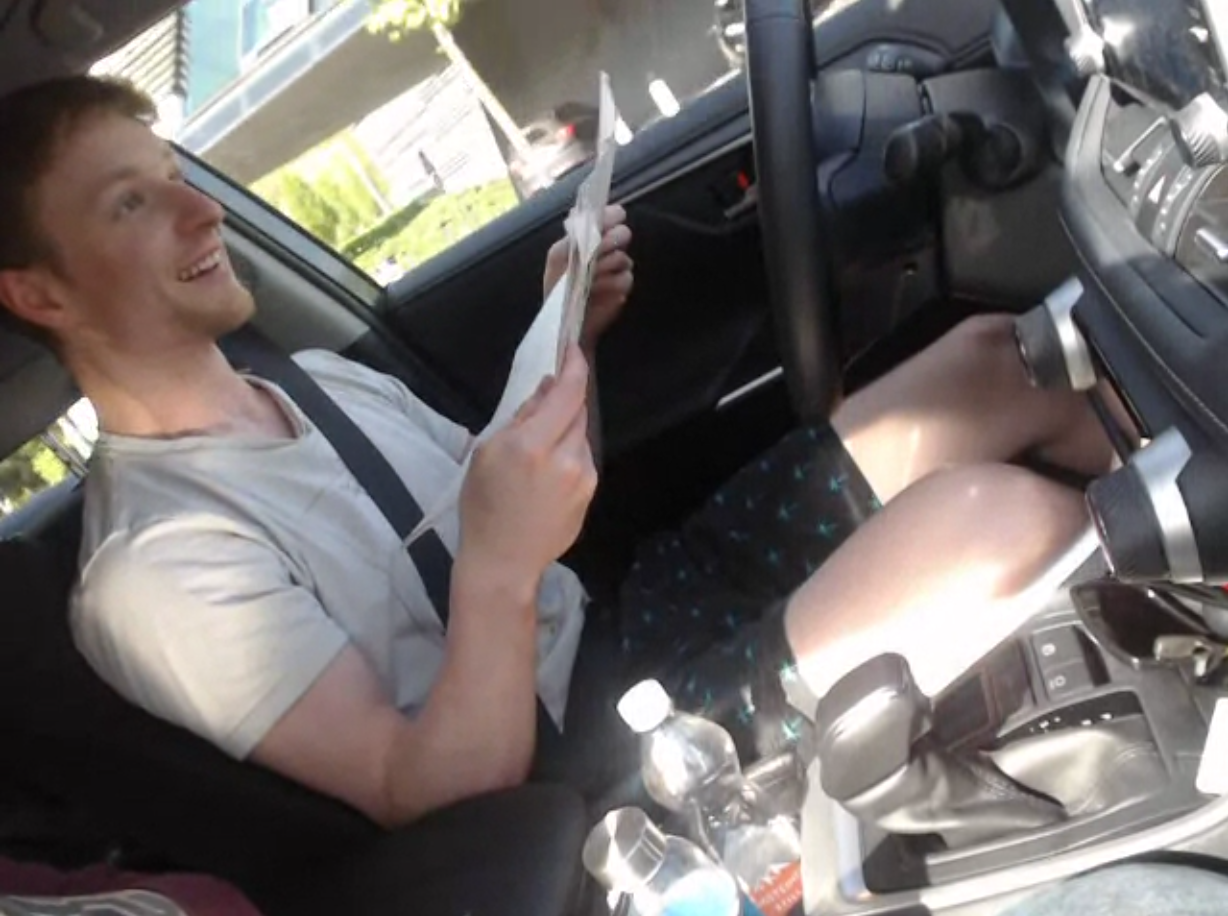}
  \caption{An image from the JKU-ITS AVDM dataset, showcasing a realistic driver reaction. Despite being tasked with reading a newspaper, the driver periodically monitors their surroundings to avoid hazardous situations.}
  \label{fig:realistic_behaviour}
\end{figure}

The JKU-ITS AVDM dataset was collected on a test road, providing authentic real-world conditions and natural driver reaction behaviors, as illustrated in Figure \ref{fig:realistic_behaviour}. This setup enhances the ability of a driver activity monitoring DL model to generalize across diverse drivers and environments. Table \ref{tab:datasetComparision} demonstrates the comparative characteristics of the real-world JKU-ITS AVDM dataset with other datasets.

\begin{table*}[htp!]

\caption{Comparison of publicly available datasets for driving-related action recognition with the proposed AVDM dataset}
  \label{tab:datasetComparision}
\centering
\begin{adjustbox}{max width=0.9\textwidth}
\begin{tabular}{|c|c|c|c|c|c|c|c| >{\bfseries}c| }
\hline
                      & HEH\cite{ohn2014hand}   & Brain4Cars \cite{jain2015car} & AUC-D.D. \cite{abouelnaga2017real}    & Drive\&Act \cite{martin2019drive} &  DMD~\cite{jOrtega2020}  &  \textbf{JKU-ITS AVDM} \\ \hline
Year                  & 2014                    & 2015                          & $2017 / 18$                           & 2019                              &  2020          &  2024         \\ \hline
Publicly Available    & $\checkmark$            & $\checkmark$                  & $\checkmark$                          & $\checkmark$                      & $\checkmark$   & $\checkmark$ \\ \hline
RGB / GrayScale       & $\checkmark$            & $\checkmark$                  & $\checkmark$                          & $\checkmark$                      & $\checkmark$   & $\checkmark$ \\ \hline
Video                 & $\checkmark$            & $\checkmark$                  & N/A                                   & $\checkmark$                      & $\checkmark$   & $\checkmark$ \\ \hline
No. Images            & N/A                     & $2 \mathrm{M}$                & $17 \mathrm{~K}$                      & $>9.6 \mathrm{M}$                 & $>39 \mathrm{~M}$ & $335 \mathrm{~K}$  \\ \hline
Resolution            & $640 \times 480$        & $1920 \times 1088$            &  $1920 \times 1080$                   & $1280 \times 1024$                & $1920 \times 1080$ & $640 \times 480$  \\ \hline
No.Subjects           & 8                       & 10                            & 31                                    & 15                                & 37             & 17          \\ \hline
Female / Male         & $1 / 7$                 & N/A                           & $9 / 22$                              & $4 / 11$                          & $10 / 27$       & $7 / 10$  \\ \hline
No. Labels            & 19                      & 5                             & 10                                    & 83                                &  93             &  8          \\ \hline
No. NDRTs             & 1                       & 0                             & 9                                     & 37                                &  13             &  7         \\ \hline
Continuous labels     & -                       & -                             & N/A                                   & $\checkmark$                      & $\checkmark$   & $\checkmark$ \\ \hline
Manual Driving        & $\checkmark$            & $\checkmark$                  & $\checkmark$                          & $\checkmark$                      & $\checkmark$   & $\checkmark$ \\ \hline
Autonomous Driving    & -                       & -                             & -                                     & $\checkmark$                      &   -            & $\checkmark$  \\ \hline
Real Varying Illumination Conditions &  $\checkmark$   & $\checkmark$           & $\checkmark$                          & -                                 & $\checkmark$   & $\checkmark$ \\ \hline
Environment           & Real Dynamic            & Real Dynamic                  & Real Static                           & Simulation                        & Real Static/Dynamic, Simulation   & Real Dynamic \\ \hline
\end{tabular}
\end{adjustbox}

\end{table*}

\begin{figure}[]
\centering
  \includegraphics[width=0.4\columnwidth]{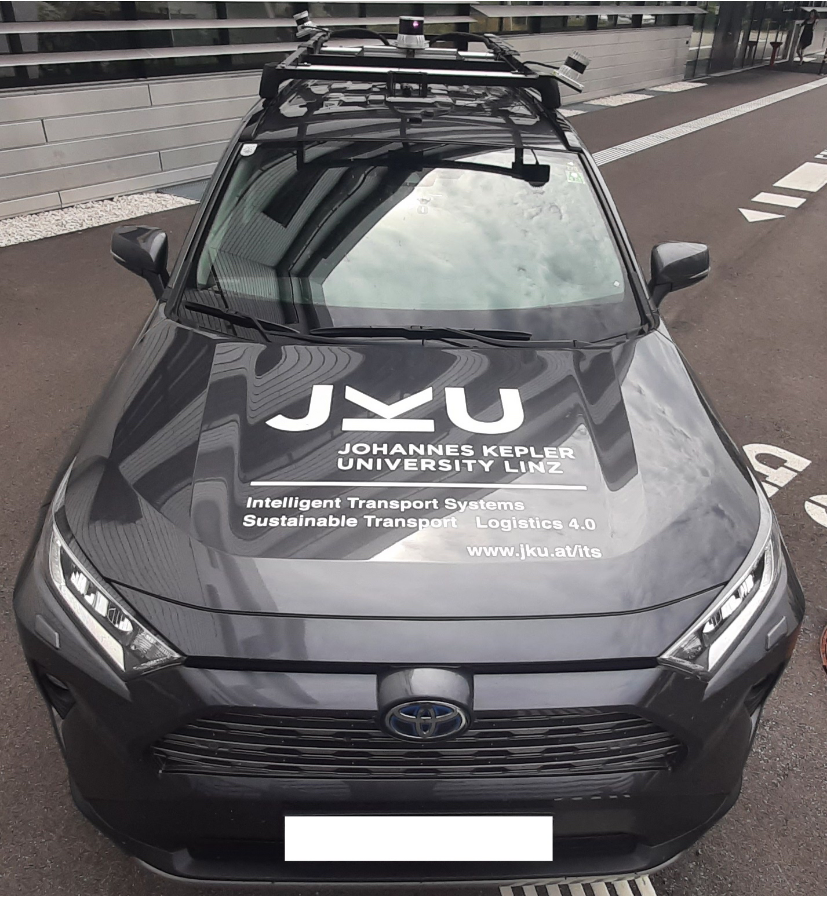}
  \caption{JKU-ITS research vehicle employed for the data collection.}
  \label{fig:JKU_its_Car}
\end{figure}

\section{The Automated Vehicle Driver Monitoring Dataset} 
\label{sec:validatingSimulation}

The following section presents the proposed JKU-ITS AVDM dataset as well as the DL network used for the driver activity classification.

\subsection{Apparatus}
The data was collected using an i7-11800H Laptop with 32 GB RAM, and a RTX 30 Series GPU on an Ubuntu 20.04 OS. For the training, a PC with an i9-9900 CPU, 32 GB RAM and an RTX 3090 was utilized.
The videos were recorded onboard the JKU-ITS research vehicle (see Figure \ref{fig:JKU_its_Car}) using a Logitech C920 webcam, which was positioned on the A-pillar of the passenger door, capturing the full driver position. This angle enabled capturing the entire motion space of driver activities, along with the objects participants interacted with during the experiment.

The participants initiated the process by activating the automated system of the vehicle, and starting the task to be performed.

The drive-by-wire system employed for control during the experiment comprised two primary components as described in~\cite{barrio2023development}: firstly, a drive-by-wire mechanism, implemented using Openpilot algorithms \cite{gHoltz}, wherein the Black Panda device transferred the acceleration and steering control to the vehicle via the internal built-in Advanced Driver Assistance System (ADAS), through a ROS Wrapper. Secondly, a custom ROS2 high-level controller for the vehicle that generated trajectories, speed profiles, and steering and acceleration commands based on pre-recorded waypoints obtained through the vehicle’s GPS. 

% the driver is the person behind the steering wheel, any place else, then refer to the human as person
To ensure safety during the experiment, a person was present in the passenger seat at all times, overseeing the system operation. In case of any malfunction / unexpected behaviour, the safety person was in charge of overriding the automated control of the vehicle using a joystick to apply the brakes. 

\subsection{AVDM Dataset Characteristics}
\label{theDataset}

\begin{table}[h]
\centering
\caption{Activities performed during the data collection}
\label{tab:activities}
\begin{tabularx}{\columnwidth}{X}
\hline
Manual Driving \\ \hline
Sitting Still in the Driver Seat \\ \hline
Using a Mobile Phone for Browsing the Internet, Texting, etc. \\ \hline
Talking on the Phone: Initiating or Replying to a Call with Another Person \\ \hline
Reading a Magazine \\ \hline
Reading a Book \\ \hline
Reading a Newspaper \\ \hline
Drinking a Beverage from a Bottle \\ \hline
\end{tabularx}
\end{table}

\begin{table}[]
\centering
\caption{Dataset classes distribution}
\label{tab:datasetDist}
\begin{tabularx}{\columnwidth}{XX}
%\begin{tabular}{|c|c|}
\hline
Class Label        & Dataset Percentage (\%) \\ \hline
talking\_phone     & 12.7                    \\ \hline
using\_phone       & 14.8                    \\ \hline
sitting\_still     & 11.2                    \\ \hline
driving            & 16.8                    \\ \hline
drinking\_bottle   & 13.2                    \\ \hline
reading\_book      & 9.7                     \\ \hline
reading\_magazine  & 9.4                     \\ \hline
reading\_newspaper & 12.3                    \\ \hline
\end{tabularx}
\end{table}
% \vspace{-10em}
The AVDM dataset consisted of a sample of 17 participants recruited from the Johannes Kepler University. They were asked to perform the 8 activities listed in Table~\ref{tab:activities}, which included manual driving and 7 non-driving-related tasks, while the vehicle autonomously navigated along a specified test route within the campus premises.
The duration of the aforementioned recorded data was of 200 minutes in the form of RGB videos with their respective RGB image folders.
Data collection occurred at various times throughout the day and under diverse weather conditions to ensure a wide range of illumination and weather scenarios. Table~\ref{tab:participant_summary} summarizes the time of day, temperature range, and weather conditions for each participant.

\begin{figure*}[htp!]
\centering
  \includegraphics[width=0.79\textwidth]{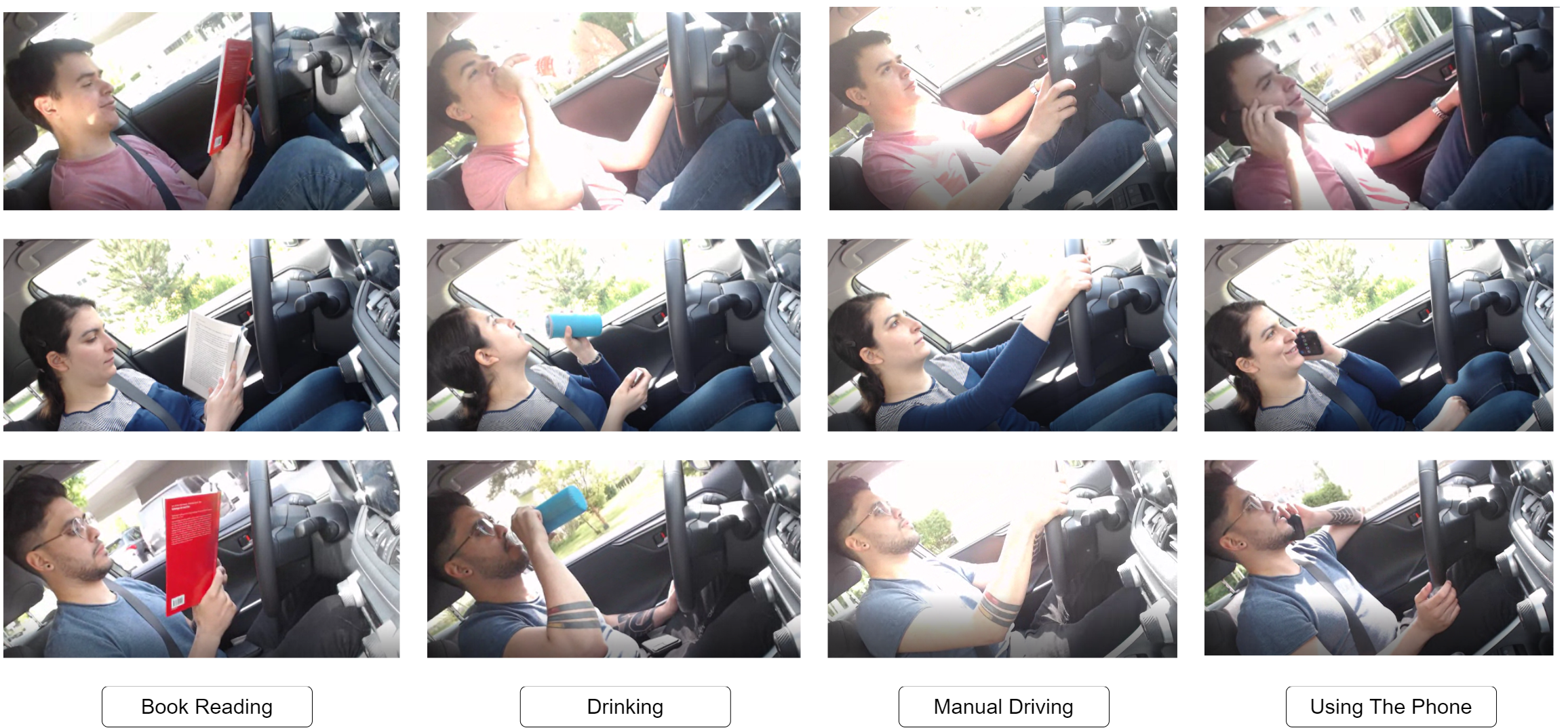}
  \caption{Image examples from the JKU-ITS AVDM dataset showcasing a variety of activities captured under varying illumination conditions.}
  \label{fig:JKU_Dataset}
\end{figure*}

\renewcommand{\tabcolsep}{3pt}
\begin{table}[h]
\centering
\caption{Summary of Participant Information}
\label{tab:participant_summary}
\begin{tabular}{|c|c|c|c|}

\hline
\textbf{Time of Day}                                         & \textbf{Temp(°C)} & \textbf{Weather} & \textbf{Participant} \\ \hline
\begin{tabular}[c]{@{}c@{}} Late Morning\\ (9-11:59 AM)\end{tabular}  & 4-14              & Cloudy      & 8-11, 14-17                 \\ \hline
\begin{tabular}[c]{@{}c@{}}Early Afternoon\\ (12 PM - 2:59 PM)\end{tabular} & 10-16 & Semi cloudy & 4,5,12,13 \\ \hline
\begin{tabular}[c]{@{}c@{}}Late Afternoon\\ (3-6 PM)\end{tabular} & 9-19             & Sunny/Semi cloudy      & 1-3,6,7                \\ \hline
\end{tabular}
\end{table}

Imbalance among classes within datasets can significantly degrade the performance of classification DL networks during training. Hence, the proportions of each class within the AVDM dataset were meticulously examined, as outlined in Table \ref{tab:datasetDist}.  The table shows that the classes are evenly distributed. This balance promotes improved training outcomes and enhances overall performance robustness of the DL network.

The data consists of two sets of labels defined as follows and exemplarized below:\\

1. Following the Charades dataset format~\cite{sigurdsson2016hollywood} to provide detailed information about each action instance in the video: 

\begin{verbatim}
"s01v01": {
"subset": "training",
"duration": 76.59699988365173,
"actions":[[0, 0, 76.59699988365173]]}
\end{verbatim}

2. Per-Frame Labelling as provided in the csv files:
\vspace{-0.5em}

\begin{verbatim}
Frame,	 Timestamp,	    Action
  0,	   1683204895,	sitting_still
  1,	   1683204895,	sitting_still
  .....
\end{verbatim}

The format of the dataset and its structure is shown below:
\dirtree{%
.1 Root Data Directory.
    .2 $s{Participant No. 1}$ .
        .3 $s{Participant No. 1}v{Video No. 01}.webm$.
        .3 $s{Participant No. 1}v{Video No. 01}.csv$.
        .3 $s{Participant No. 1}v{Video No. 01}$.
        .3 ....
        .3 $s{Participant No. 1}v{Video No. N}.webm$.
        .3 $s{Participant No. 1}v{Video No. N}.csv$.
        .3 $s{Participant No. 1}v{Video No. N}$.
    .2 ....
    .2 $s{Participant No. 17}$.
        .3 $s{Participant No. 17}v{Video No. 01}.webm$.
        .3 $s{Participant No. 17}v{Video No. 01}.csv$.
        .3 $s{Participant No. 17}v{Video No. 01}$.
        .3 $....$ .
        .3 $s{Participant No. 17}v{Video No. N}.webm$.
        .3 $s{Participant No. 17}v{Video No. N}.csv$.
        .3 $s{Participant No. 17}v{Video No. N}$.
    .2 $actions.names$.
    .2 $labels.json$.
}

All the videos within the dataset are in webm format. Each video has its respective folder, with the extracted frames in png format. The class labels can be found in $actions.names$ and the labels of all the videos can be found in $labels.json$. There are label files for each video file, which is in the csv format. This csv file contains a label for each frame.

\begin{figure*}[ht]  
\centering
\includegraphics[width=0.75\textwidth]{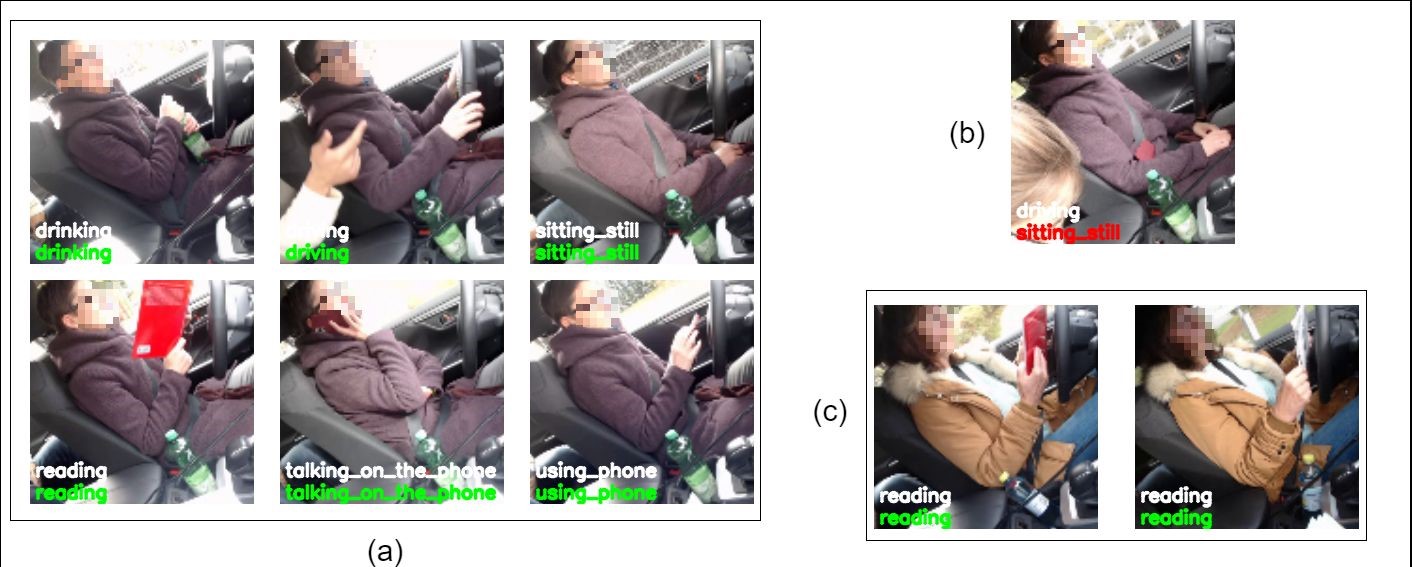}
  \caption{Exemplary images from the AVDM dataset for predicting actions using the I3D model with the 6 labelled classes version of the AVDM dataset. Labels in the bottom left corner of each image are color-coded as: white for ground truth, green for correct classifications, and red for incorrect classifications. In (a), we showcase instances from the test set where the network accurately predicted the actions; (b) illustrates a challenging scenario where distractions in the environment pose difficulties for the DL network: the proximity of the hands to the steering wheel might mislead the network into predicting driving instead of sitting still; (c) presents a test sample depicting the similarity between two closely related classes: reading a book and reading a magazine. Due to the resemblance in posture and the items being used, the network finds it easier to detect the general activity rather than classifying the specific type of object being read.}
\label{fig:results_I3D}
\end{figure*}

\subsection{Labeling Process}
\label{validation}
\begin{figure}[!h]
  \centering
  \includegraphics[width=0.8\columnwidth]{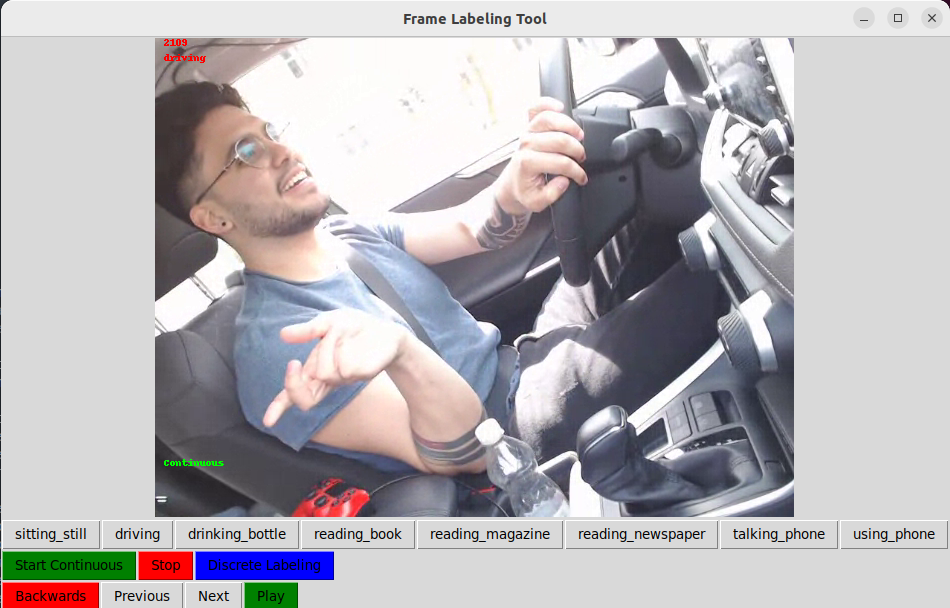}
  \caption{The labeling tool developed accepts the video path as input, allowing users to navigate through individual frames of the video. Users can easily switch between assigned labels for each frame and utilize the tool to automatically label frames with the selected label.} 
  \label{fig:labelling tool}
\end{figure}

To label the dataset, a semi-automatic labelling tool was developed. This tool takes in a video as an input, then prompts the user which label to select for the frames. When a label class is selected, the user is given the option to have the tool automatically label the subsequent frames with the selected label.
% the user is prompted the option to make the tool automatically label the next frames with the selected label. 
If the user finds a frame / some frames mislabeled, there is an option to manually go back to these frames, change the label, then resume the automatic labelling. After going over each video, the tool outputs a csv file with the labels associated with each frame of the video. The interface of the built tool can be seen in Figure \ref{fig:labelling tool}.

Annotations were provided in two formats and included two sets of classes for the Charades format: a 6-class version, which included driving, sitting still, using the phone, talking on the phone, reading, and drinking. In this version, all reading-related classes were merged into a single category. The second format was an 8-class version, where the reading class was split into reading a book, newspaper, and magazine.

\subsection{Baseline Model} 
The primary objective of the model was to accurately predict every action performed by each participant, utilizing a minimum of 64 frames. For this task, we trained the I3D Model from \cite{carreira2017quo} with the JKU-ITS AVDM dataset.
The I3D adapts the 2D filters of Inception-v1 into a temporal dimension and processes 64-frame video snippets of 224x224 resolution. The network consists of 27 layers, with nine Inception modules executing parallel convolutions and concatenating the output, which in turn, enhances the computational efficiency.

The training and evaluation process were performed on the dataset, utilizing Charades-format labels with the 6 primary classes described above (see Figure~\ref{fig:results_I3D}). 

The I3D network was trained using data from 15 out of the 17 participants and then tested using data from the remaining 2 participants. The network was trained for 64,000 steps with a batch size of 20, a learning rate of 0.1 (decreasing to 0.01 at 300 steps and to 0.001 at 1000 steps) and a Momentum of 0.9.

\begin{figure}[!]
  \centering
  \includegraphics[width=0.85\columnwidth]{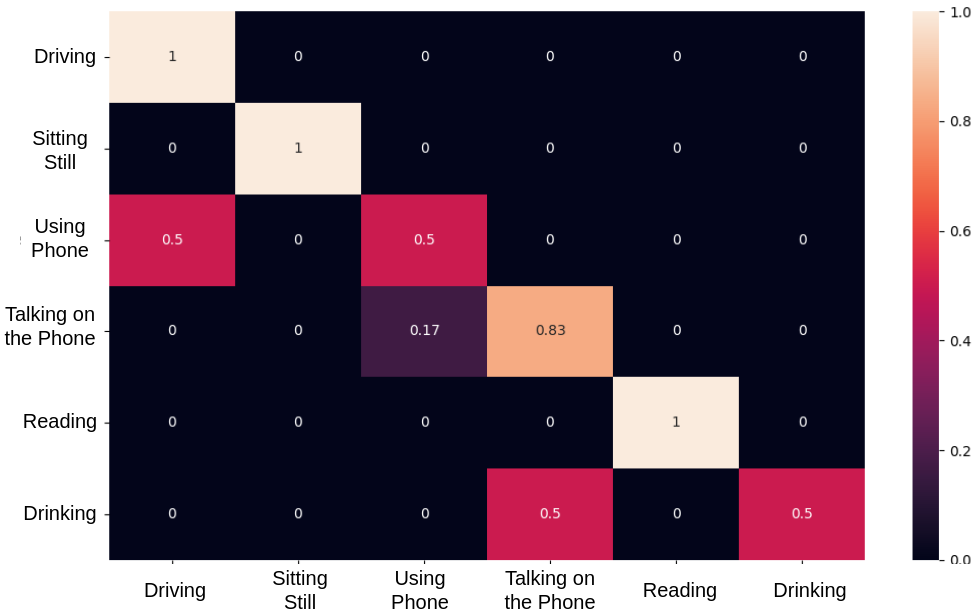}
  \caption{ Confusion matrix generated on the test set representing the distribution of predictions across the 6 classes.}
  \label{fig:Confusion martix}
\end{figure}

\section{ Training Results }
\label{sec:ExperimentalResults}

% \subsection{Results}

\subsection{Qualitative results}
% TODO reorder the figures to have an asending numbering

The results of the presented approach showed the network's strong performance and clear action distinction, as depicted in Figure \ref{fig:results_I3D}, even with training data characterized by varying illumination and weather conditions.

After analyzing each class individually, any action that did not involve a small object that can be mostly occluded within the 6-class label version (as described in Section~\ref{validation}) was accurately identified. 
However, in the context of the "drinking" class, as the bottle becomes mostly occluded in some cases, as well as having the hands very close to the position of making a phone call, this class is mixed with "talking\_on\_the\_phone". In case of "using\_phone", the phone itself becomes mostly occluded in cases. The addition of having the hands very close to the steering wheel causes the model to confuse this class with the driving class.

Findings from the 8-class version, which included various interactions with a phone and reading different materials, showed that the model typically remained within the reading category, even if, for example, a video clip depicting a driver reading a magazine was inaccurately categorized by the network as reading a book (see Figure \ref{fig:results_I3D}(c)).

\subsection{Quantitative results}

The outcomes obtained from running the I3D model on the test set showed that the model classifies actions with high accuracy. Figure \ref{fig:Confusion martix}
shows the confusion matrix. The results indicate a 100\% accuracy for the driving, sitting\_still and reading classes, 83\% for talking on the phone and 50\% for both drinking and using the phone. The model incorrectly classified certain "using phone" frames as "driving" due to the aforementioned closeness of the test subjects' hands to the steering wheel as well as occlusion in specific scenarios. For the drinking class, the water bottle becomes close to the face, near a location of a cell phone when calling someone, thus explaining the DL network mix up.

\section{Conclusion and Future Work}
\label{sec:Conclusions}

This paper introduced a novel dataset collected from real-world automated driving trials, providing a benchmark for implementing models for driver activity recognition.

The dataset comprises data from 17 participants engaged in various activities during automated navigation. Annotations were provided in two formats, offering two sets of classes.
To set a benchmark score on the dataset, we trained and tested the I3D DL network. 

Future research will explore expanding the dataset with additional classes and subclasses to enhance the model's classification capabilities. Modifications to the vehicle cockpit will be implemented to diversify the dataset and improve generalization. Efforts will be made to increase input data variation, aiming to achieve more robust applications. Furthermore, the real-world applicability of the dataset will be enhanced by incorporating interior sensing systems with higher dynamic range.

\section{Acknowledgment}
This work was partially supported by the Austrian Science Fund (FWF), project number P 34485-N and by the Austrian Research Promotion Agency (FFG) project number: 48652784.
%%%%%%%%%%%%%%%%%%%%%%%%%%%%%%%%%%%%%%%%%%%%%%%%%%%%%%%%%%%%%%%%%%%%%%%%%%%%%%%%

\bibliographystyle{IEEEtran}
\bibliography{paper}

\end{document}